\documentclass[letterpaper]{article} 
\usepackage{aaai25}  
\usepackage{times}  
\usepackage{helvet}  
\usepackage{courier}  
\usepackage[hyphens]{url}  
\usepackage{graphicx} 
\usepackage{natbib}  
\usepackage{caption} 
\usepackage{algorithm}
\usepackage{algpseudocode}

\usepackage{newfloat}
\usepackage{listings}
\DeclareCaptionStyle{ruled}{labelfont=normalfont,labelsep=colon,strut=off} 
\floatstyle{ruled}
\newfloat{listing}{tb}{lst}{}
\floatname{listing}{Listing}
\title{Efficient Neural Network Encoding for 3D Color Lookup Tables}
\author{
   Vahid Zehtab\thanks{Work done while an intern at the Samsung AI Center Toronto.}\textsuperscript{\rm 1, 2, 3}, 
    David B. Lindell\textsuperscript{\rm 1, 2},
    Marcus A. Brubaker\textsuperscript{\rm 1, 2, 3, 4},
    Michael S. Brown\textsuperscript{\rm 3, 4}
}
\affiliations{
    \textsuperscript{\rm 1}University of Toronto,
    \textsuperscript{\rm 2}Vector Institute,
    \textsuperscript{\rm 3}Samsung AI Center Toronto,
    \textsuperscript{\rm 4}York University,
    \\

    \{zehtab, lindell\}@cs.toronto.edu,\ \ \{mab, mbrown\}@eecs.yorku.ca
}

\usepackage{placeins} 
\usepackage{bm}
\usepackage[thinc]{esdiff}
\usepackage{multirow}
\usepackage{amsmath} 
\usepackage{amssymb} 
\usepackage{subcaption}
\usepackage{arydshln}
\usepackage[dvipsnames]{xcolor}

\DeclareMathOperator*{\expected}{\mathbb{E}}
\DeclareMathOperator*{\argmin}{argmin}
\DeclareMathOperator{\lipswish}{LipSwish}

\newcommand{\idx}{\mathbf{o}}
\newcommand{\Idxs}{\mathcal{L}}
\newcommand{\embedmat}{E}

\newcommand{\makesupplementtitle}[1]{%
  \twocolumn[ 
    \centering
    \vskip 2.0em 
    {\LARGE \bfseries #1 \par} 
    \vskip 4.0em 
  ]
}

\begin{document}

\maketitle

\begin{abstract}
3D color lookup tables (LUTs) enable precise color manipulation by mapping input RGB values to specific output RGB values. 3D LUTs are instrumental in various applications, including video editing, in-camera processing, photographic filters, computer graphics, and color processing for displays.  While an individual LUT does not incur a high memory overhead, software and devices may need to store dozens to hundreds of LUTs that can take over 100 MB.  This work aims to develop a neural network architecture that can encode hundreds of LUTs in a single compact representation. To this end, we propose a model with a memory footprint of less than 0.25 MB that can reconstruct 512 LUTs with only minor color distortion ($\bar{\Delta}E_M$ $\leq$ 2.0) over the entire color gamut.  We also show that our network can weight colors to provide further quality gains on natural image colors ($\bar{\Delta}{E}_M$ $\leq$ 1.0). Finally, we show that minor modifications to the network architecture enable a bijective encoding that produces LUTs that are invertible, allowing for reverse color processing. Our code is available at \url{https://github.com/vahidzee/ennelut}.
\end{abstract}

\section{Introduction}

\label{sec:intro}
Color manipulation is a fundamental operation in computer vision and image processing, where input RGB values map to output RGB values. A widely used method for encoding such mappings is through a 3D color lookup table (LUT). LUTs are employed in a diverse range of applications, such as video editing, in-camera processing, photographic filters, computer graphics, and color processing for displays. Particularly, LUTs play a pivotal role in ensuring color accuracy and consistency across various display hardware.

An individual 3D LUT imposes a manageable memory overhead. For example, a standard 
33$\times$33$\times$33
LUT at 16-bit precision requires approximately 70 KB. However, professional LUTs for color grading or color management often rely on a 
65$\times$65$\times$65
resolution that requires approximately 0.5 MB at 16-bit precision. Storing a library of hundreds of such LUTs can quickly become a limitation, especially for applications running on resource-constrained devices, such as smartphones and camera hardware. Black-box compression (e.g., zip) provides limited reduction, for instance, a library of 512 zipped LUTs requires approximately 124 MB.

\begin{figure}[t]
    \centering
    \includegraphics[width=1.0\linewidth]{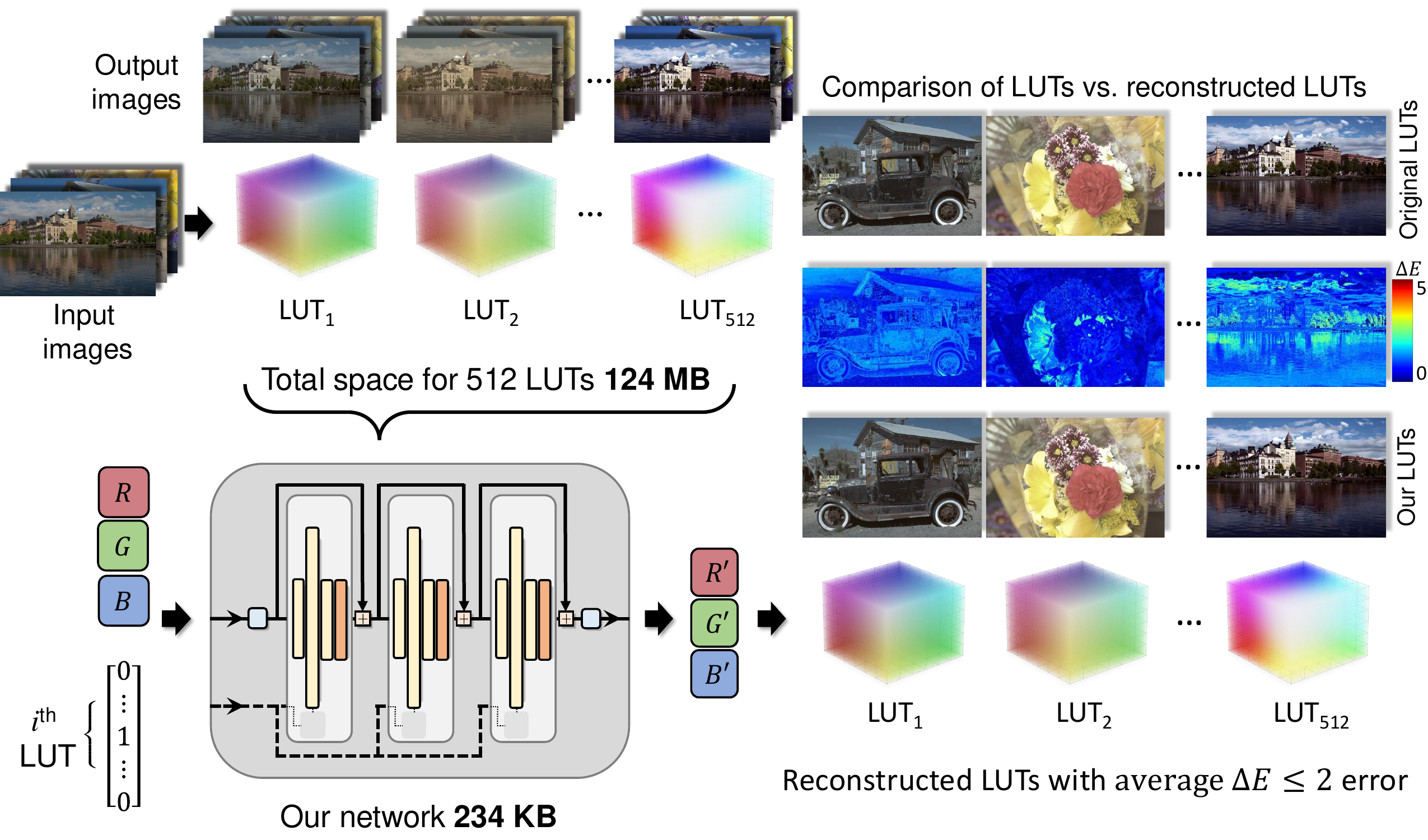}
    \caption{We propose a neural network architecture that encodes hundreds of LUTs into a single representation at a fraction of the memory requirements.  Encoded LUTs can be reconstructed with minimal color distortion~(i.e., $\Delta E\leq2$).}
    \label{fig:teaser}
\end{figure}

In this work, we propose a compact neural representation capable of reconstructing individual LUTs on the fly, reducing storage requirements and enhancing real-time color manipulation capabilities.  We examine four network size variants
for embedding different numbers of LUTs, with the largest requiring only 0.3 Mb storage to encode 512 LUTs.
Our model recovers a full-size LUT in under 2 milliseconds, accommodating real-time applications. The recovered LUTs incur an average color difference ($\Delta E$) of less than or equal to 2.0, an industry-standard definition for acceptable perceptual color distortion~\cite{sharma2017digital}.

We demonstrate the versatility of our proposed neural network by introducing alternative loss functions and a straightforward weighting of input colors, enabling enhanced quality improvements when focused on natural image color gamuts. Our alternative weighted training method achieves color differences ($\Delta E$) of less than $1.0$ for natural images.

Lastly, we explore a modification to the network architecture, allowing for a bijective encoding of LUTs. This bijective encoding makes it possible to produce LUTs that are inherently invertible, opening up opportunities for inverse color processing and expanding the utility of 3D LUTs in computer vision and image processing.

\section{Related Work}
\label{related_work}
\subsection{LUTs for Color Manipulation}
As non-parametric function approximators, 3D Color LUTs (CLUTs or simply LUTs) are suitable for modeling complex transformations by sampling a target transformation on a 3D lattice.
LUTs are instrumental for color grading in video editing~\cite{ColorGrading} and used for correction in display colors~\cite{ColorDisplays}.
LUTs are also essential tools in various stages of ISP pipelines~\cite{kasson1995performing,karaimer2018improving}, where they help ensure colorimetric accuracy or are used to render different picture styles~\cite{karaimer2016software, delbracio2021mobile,zhang2022clut}.  

LUTs have also been used as building blocks in frameworks that learn image enhancements~\cite{zeng2020learning,wang2021real}.
For instance, work in \cite{yang2022adaint} learned LUTs using an adaptive lattice for color manipulation, while \cite{wang2021real,liu20234d} learn contextualized LUTs for spatially varying and image-dependant color transformations. 
Although such works typically deal with less than a handful of LUTs, the memory footprint of LUTs has enticed attempts at less memory-intensive formulations such as through a combination of 1D and 3D LUTs~\cite{yang2022seplut}, or the decomposition of 3D LUTs into learned sub-tables and lower-rank matrices~\cite{zhao2022learning}.
In this work, we assume the LUTs are provided and seek a neural architecture to encode them efficiently.

\subsection{LUT Compression} 
LUTs are stored as a 3D input-output lattice, where the input lattice is usually set to a uniform grid at fixed intervals along the R, G, and B color axes.
As a result, it is necessary only to store the output colors.
LUTs are commonly stored as standard ASCII or Unicode (i.e., \texttt{.cube} files) or as a binary array of floating-point values.
Such encodings can be further compressed using off-the-shelf compression algorithms (e.g., zip).
Alternatively, a LUT can be stored as an RGB image called a Hald image.
The Hald image resolution can be adjusted to mimic different LUT resolutions.
Hald images are stored in a lossless format such as \texttt{png}.
Hald images also serve to visualize how an individual LUT manipulates colors over the entire color space.  
    
Various lossless compression techniques have been proposed~\cite{balaji2007hierarchical,balaji2008preprocessing, shaw2012lossless}, achieving average compression rates of $\approx$30\% (similar to the compression rates of \texttt{png}).
In \cite{tang2016icc}, lossy compressed LUTs were stored as ICC device-link profiles.
In \cite{tschumperle20193d,tschumperle2020reconstruction}, sparse color key points were estimated that enabled the reconstruction of the original 3D LUT, providing an average compression rate of $\ge$95\%.
Existing LUT compression methods target individual LUTs.
Our network implicitly compresses multiple LUTs, capitalizing on their inherent similarities, and achieves compression rates of $\ge$99\%.
    
\subsection{Neural LUTs} 
To our knowledge, \cite{conde2024nilut} is the only work to target LUT embedding and reconstruction using a neural network. 
Specifically, ~\cite{conde2024nilut} showed that 3 to 5 LUTs could be encoded using a model requiring approximately 750 KB storage. 
We show that our network architecture is better suited for LUT embedding.
In addition, we describe how a straightforward modification to our network imposes bijectivity on the LUTs encoding.
This allows an image processed with our LUT to be restored to its initial RGB values.

\begin{figure*}[]
    \begin{center}
        \includegraphics[width=0.95\linewidth]{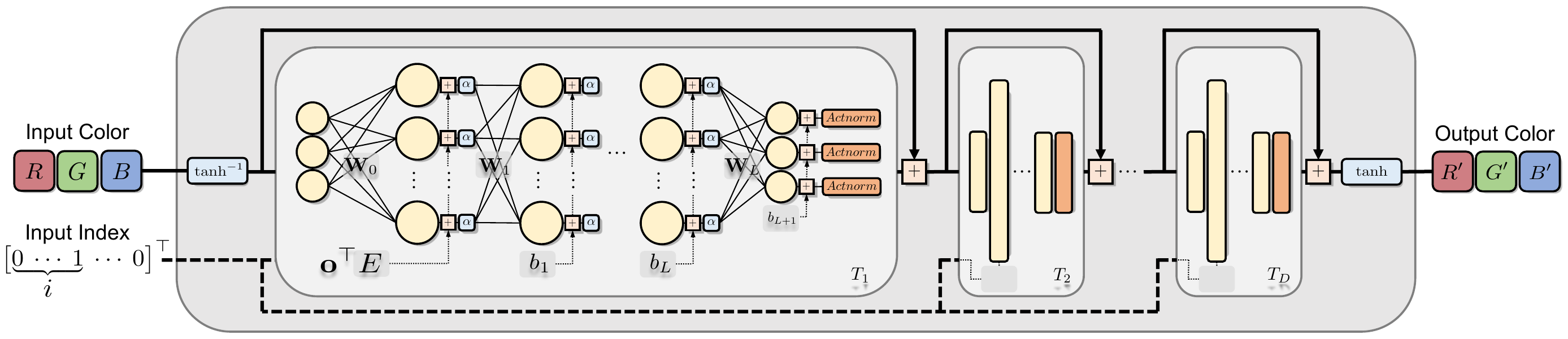}
    \end{center}
    \caption{
        Our network consists of $D$ transformations $T_i(\cdot, \idx)$, conditioned on a specific LUT by a one-hot encoded index $\idx$ indicating the desired LUT to use. $T_i$s contribute to the reconstructed output color through consecutive residual additions. $T_i$s are modeled with multilayer perceptrons (MLP) with $\alpha$ non-linearities, where the biases of their first layer are selected based on $\idx$. Activation normalization
        ~\cite{kingma2018glow}
        is used after each transformation to control the magnitude of the residual functions, ensuring stability in deeper architectures. $\tanh^{-1}$ and $\tanh$ respectively transform the inputs and outputs of the network, bringing it closer to the local identity.
    }
    \label{fig:overview}
\end{figure*}

\section{Proposed Approach}
\label{method}
Formally, a 3D color LUT $F: \mathbb{R}^3 \rightarrow \mathbb{R}^3$ is a function that maps input RGB colors to output RGB colors represented by a set of input-output pairs on a sparse lattice covering the input color space.
$F$ computes the output color of an arbitrary input using traditional interpolation techniques~\cite{kasson1995performing}. 

Given a set of LUTs $\{F_1, F_2, \cdots, F_N\}$, we aim to find an implicit neural representation $f_\theta(\cdot, \idx): \mathbb{R}^{3} \times \Idxs \rightarrow\mathbb{R}^3$, where $\Idxs \subset \mathbb{R}^{N}$ is the set of LUTs and $f$ is a function modeled with a deep neural network parameterized by $\theta$.
The function $f_\theta$ takes on an \texttt{RGB} input in addition to a desired LUT $F_i$, represented with a one-hot encoded vector $\idx_i \in \Idxs$. 

We learn the parameters $\theta$ such that $f_\theta(., \idx_i) \approx F_i(.)$.
We formalize this in terms of the optimization problem
    \begin{align}
        \theta^*=&\argmin_\theta\ \expected_{x\sim\mathcal{P}}\left[\sum_i^N \lVert f_\theta(x, \idx_i) - F_i(x)  \rVert_2^2 \right],
    \end{align}
where $\mathcal{P}$ signifies a probability distribution over the input colors.
Different choices for $\mathcal{P}$ result in different $\theta^*$s trading off better reconstruction of certain colors over the others.
The choice of $\mathcal{P}$ depends on the intended use case of $f_{\theta^*}$.
As we show in our experimental results, the closer $\mathcal{P}$ is to the evaluation distribution, the better $f_{\theta^*}$ trades off the quality of color reconstructions over out-of-distribution colors.

\subsection{Network Architecture}
To design an efficient architecture, we desire to capture the inherent characteristics of LUTs.
We aim to devise a network structure that embodies the following traits: 
\begin{enumerate}
    \item LUTs frequently resemble an identity function in many regions of the input space,
    \item the majority of LUTs exhibit local bijectivity, with exceptions being intentional manipulations that compress the color space into lower-dimensional manifolds (e.g., RGB to grayscale LUTs).
\end{enumerate}

Residual networks (ResNet)~\cite{he2016deep} are a natural fit for the first property, given their inductive bias toward an identity function. To capture the second property, we propose to utilize architectures from the normalizing flows literature~\cite{kobyzev2020normalizing}. 

Normalizing flows~\cite{rezende2015variational} are a class of bijective neural networks often used for generative modeling and density estimation.
We take inspiration from~\cite{jacobsen2018revnet,chen2019residual,behrmann2019invertible} that advocate modifying a stock residual network with an inductive bias towards bijective maps.
As demonstrated in~\cite{jacobsen2018revnet}, if all the residual functions in a ResNet have a Lipschitz constant strictly smaller than $1$ the entire network is invertible by the Banach fixed-point theorem.
For such networks, termed residual flows, each residual function can be explicitly restricted through spectral normalization~\cite{gouk2021regularisation,miyato2018spectral} and activation functions that have bounded derivatives.
However, given that not all LUTs are bijective~(e.g., RGB to greyscale), we do not want to rigidly restrict the network to bijectivity.
Instead, we initialize it close to such a bijective transformation as a form of inductive bias to regularize learning.

To this end, we design our neural architecture with $D$ residual components $T_1, T_2, \dots, T_D$, as shown in Figure~\ref{fig:overview}.
Each residual component $T_i$, is an MLP with $\lipswish$ non-linearities~\cite{chen2019residual}.
$T_i$s are conditioned on the desired LUT $\idx$ by a learned matrix $\embedmat_i\in \mathbb{R}^{N\times h}$, where $N$ is the number of embedded LUTs, and $h$ is the width of the first hidden layer in $T_i$.
The one-hot encoded vector $\idx$, selects a row from $\embedmat_i$ ($\idx^\top \embedmat_i$), which is used as the biases for the first hidden layer in $T_i$.

We use activation normalization~\cite{kingma2018glow} to mitigate numerical instabilities caused by stacking multiple residual functions together.
To ensure that our model is as close to the space of identity and bijective functions, we initialize all the weights and biases of the network with small values and use $\lipswish$ non-linearities with bounded derivatives.

Moreover, to deal with the bounded normalized input-output space, we convert the inputs using $\tanh^{-1}$ and then return to the bounded color space by applying $\tanh$ on the network outputs.
This way, the learnable part of the model always operates in an unbounded space with similar input-output measures.
This makes for a better computational model and helps our model deal with colors with high saturation, as the network weights no longer need to grow to produce saturated colors.
In addition, compared to clipping the values as in \cite{conde2024nilut}, our method does not suffer from clipped gradient values, which can hinder training.

\subsection{Training}
Our training accommodates the simultaneous embedding of many LUTs on a custom color distribution $\mathcal{P}$.
At each optimization iteration, we perform the following:
\begin{enumerate}
    \item Randomly pick a batch of input colors from the $256^3$ input color space based on $\mathcal{P}$ and normalize the source color values.
    \item Compute the target colors by applying all (or optionally a random subset of LUTs) on the batch and normalize the target color values.
    \item Compute the $L_2$ reconstruction error and update the weights. The network can also be trained using a $\Delta E$ loss which is described in our alternative training approaches experiments section.
\end{enumerate}
\noindent\textbf{Normalization.} We normalize each color channel to fall within the range $[-0.83, 0.83]$ ensuring that the values passed through $\tanh$ and $\tanh^{-1}$ remain in a region with sufficiently large gradients, reducing the risk of vanishing gradients. Refer to supplemental materials for details.

\begin{table*}[t]
    \centering
    \resizebox{0.8\linewidth}{!}{%
        \begin{tabular}{l:c|ccc|ccc}
            \multirow{2}{*}{Model (size)}	& Compression & \multicolumn{3}{c}{Evaluation on a $256^3$-Hald image} & \multicolumn{3}{c}{Evaluation on natural images} \\
             & ratio ($\%\uparrow$) & $\bar{\Delta} E_{M}$$\downarrow$ & $\bar{\Delta} E_{90\%}$$\downarrow$ & PSNR (dB)$\uparrow$ & $\bar{\Delta} E_{M}$$\downarrow$ & $\bar{\Delta} E_{90\%}$$\downarrow$ & PSNR (dB)$\uparrow$ \\
            \hline
            & & & & & & & \\[-7pt]
            Tiny (79 KB) & 99.94 & $4.47{\scriptstyle \pm0.04}$ & $8.61{\scriptstyle \pm0.08}$ & $32.37{\scriptstyle \pm0.09}$
            & $4.76{\scriptstyle \pm0.03}$ & $8.23{\scriptstyle \pm0.06}$ & $33.77{\scriptstyle \pm0.19}$\\
            & & & & & & & \\[-7pt]
            Small (157 KB) & 99.87 & $2.69{\scriptstyle \pm0.03}$ & $5.26{\scriptstyle \pm0.07}$ & $36.70{\scriptstyle \pm0.10}$ & $3.10{\scriptstyle \pm0.04}$ & $5.49{\scriptstyle \pm0.07}$ & $37.52{\scriptstyle \pm0.10}$\\
            & & & & & & & \\[-7pt]
            Medium (235 KB) & 99.81 & $2.00{\scriptstyle \pm0.02}$ & $3.91{\scriptstyle \pm0.04}$ & $39.31{\scriptstyle \pm0.08}$ & $2.37{\scriptstyle \pm0.02}$ & $4.24{\scriptstyle \pm0.04}$ & $39.80{\scriptstyle \pm0.07}$\\
            & & & & & & & \\[-7pt]
            Large (313 KB) & 99.75  &  $1.64{\scriptstyle \pm0.01}$ & $3.23{\scriptstyle \pm0.03}$ & $41.03{\scriptstyle \pm0.09}$ & $1.98{\scriptstyle \pm0.02}$ & $3.60{\scriptstyle \pm0.03}$ & $41.37{\scriptstyle \pm0.09}$\\
        \end{tabular}
    }
    \vspace{0.2cm}
    \caption{This table reports the quality of our reconstructed LUTs for each model size when embedding 512 LUTs and trained using uniform sampling over the color space and an $L_2$ training objective in RGB. The results are computed for Hald images and natural images. Results are averaged over ten trial runs. We refer to the compressed file size of the model checkpoint as model size. Compression ratios compare the model size to the average compressed file size of 512 binary LUTs.}
    \label{tab:hald-512}
\end{table*}

\noindent\textbf{Implementation.} During training, input colors are processed with the LUT to produce target color values. We target all LUTs at each training step. We implemented a GPU-based trilinear interpolation using PyTorch~\cite{paszke2019pytorch}.
An Nvidia RTX4090 GPU was used for training.  
With LUT interpolation implemented on GPU, we could process batch sizes of 2048 input colors, all transformed with up to 512 different LUTs.

We optimize our network using Adam \cite{kingma2014adam} with default settings and use a stepped learning rate schedule, decreasing the learning rate at fixed intervals. See supplemental materials for more details.

\subsection{Evaluation}
To evaluate the quality of the LUT approximation, we use the $\texttt{CIE}_{76}$ $\Delta E$  metric that has a direct interpretation in terms of human perception. Specifically, two colors with a $\Delta E \leq 2$ are generally considered indistinguishable for an average observer~\cite{sharma2017digital}.
As $\Delta E$ is computed per-color pair, to get an estimate of the overall qualities of color reconstructions, we track the general statistics of $\Delta E$s over the set of evaluation colors.
We use $\Delta E_{q\%}$ and $\Delta E_{M}$ to denote the $q$-th quantile and the empirical mean of $\Delta E$s over a particular evaluation set.
Since our model embeds multiple LUTs at a time, we average such statistics over all the embedded LUTs in the model to get $\bar{\Delta} E$s.
For instance, a model with $\Bar{\Delta} E_{90\%} < 2$ can reconstruct $90\%$ of the evaluated colors with a $\Delta E <2$ on average over its embedded LUTs.
Note that we also report PSNR values to be consistent with previous work.

\begin{figure}[t]
    \centering
    \includegraphics[width=0.95\linewidth]{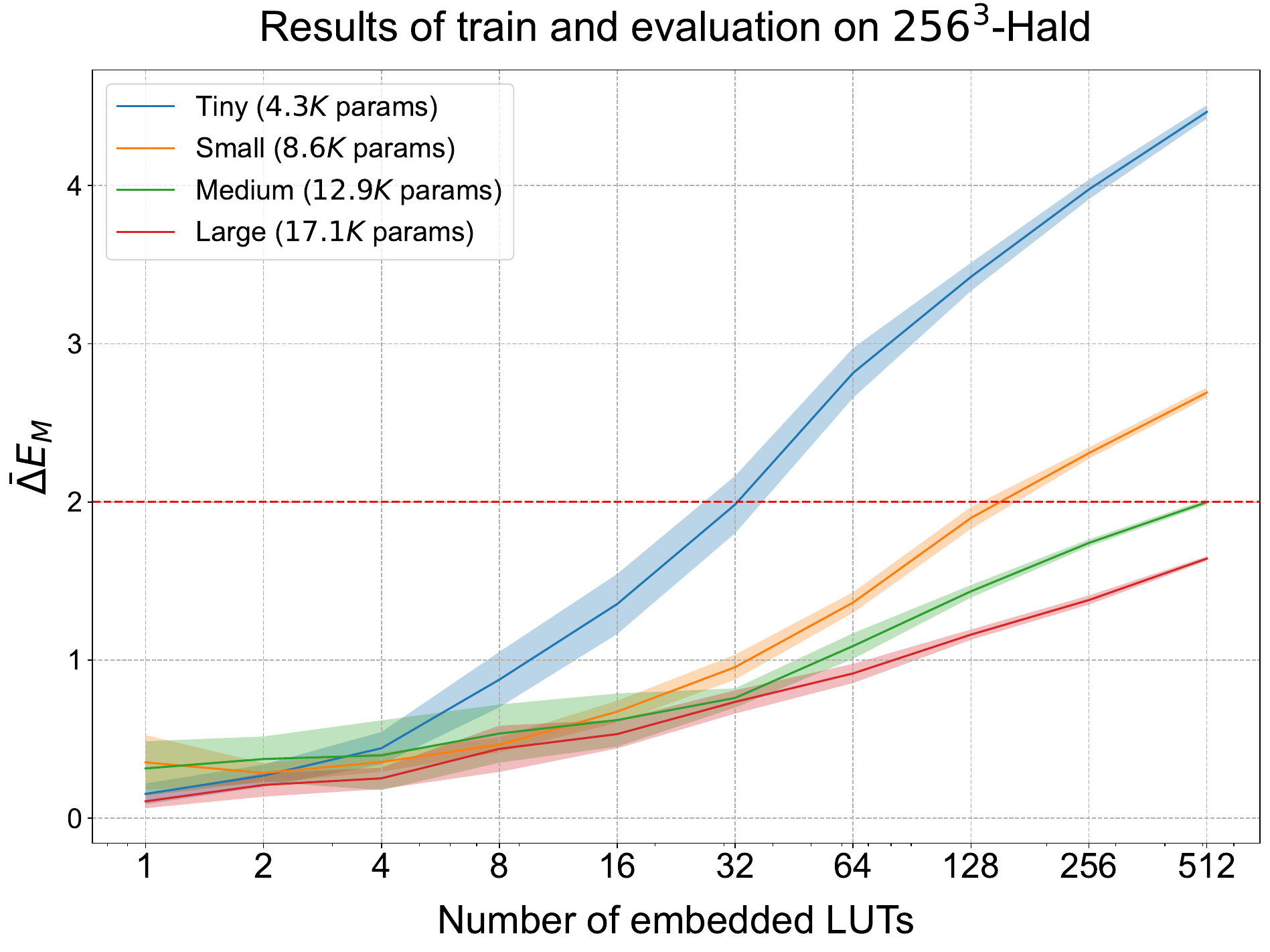}
    \caption{This plot shows how varying the number of embedded LUTs affects the performance of our models when training uniformly on the color space using an L2 loss function in RGB and evaluating against $256^3$ Hald images. The specified number of parameters represents the parameters in the $T_i$ blocks without counting the $E_i$s. Results are averaged over ten runs per \#LUTs, each using a different set of LUTs.}
    \label{fig:hald-scaling}
\end{figure}

\section{Experiments}
\label{sec:experiments}
We begin by describing our training and testing data. We evaluate our network in two scenarios: (1) uniform sampling of the color space and (2) sampling based on natural images.
Qualitative results are also shown on images processed by our reconstructed LUTs.
Finally, we demonstrate the network's ability to invert a LUT.

\subsection{Experimental Setup}
\subsubsection*{LUTs.} We obtained 543 open-source LUTs available under Creative Common licensing to train and test our method.
The LUTs range in size between $16^3$ to $35^3$ and are encoded as \texttt{.cube} files. See supplemental material for details.

\subsubsection*{Evaluation.}~We report $\Delta E$ on $256^3$ Hald images that capture all possible 8-bit color outputs produced by a given LUT.
To estimate the reconstruction quality of colors in natural images, we report the metrics averaged over 100 randomly selected images from the Adobe-MIT 5K dataset~\cite{bychkovsky2011learning}. Our experiments showed minimal differences when using all of the images from the Adobe 5K dataset~\cite{bychkovsky2011learning} for evaluation versous using a subset of 100 random selected images from the dataset. Therefore, for raster experimentation, we report the results of our evaluations on natural images using only 100 random images.
See supplemental materials for details.

The selected images were converted to 8-bit sRGB format and rescaled such that their maximum dimension is $1024$ pixels.
We ensure the reproducibility of our experiments by retraining the same model with different sets of randomly selected LUTs to provide
the average and $95\%$ confidence interval for each metric. We provide the results for different model sizes and number of embedded LUTs.

\subsubsection*{Runtime.}~Our model reconstructs LUTs at any resolution. Execution time to recover LUTs with our medium-sized model at resolutions 65$^3$, 33$^3$, 11$^3$ and 7$^3$ are $4.11$ ms, $1.64$ ms, $1.27$ ms, $1.21$ ms respectively, measured on an Nvidia RTX4090 GPU. See supplemental materials for details and additional runtime results.

\begin{table*}[t]
    \centering
    \resizebox{0.8\linewidth}{!}{%
        \begin{tabular}{c|c|ccc|ccc}
            Training & \multirow{2}{*}{Training distribution} & \multicolumn{3}{c}{Evaluation on a $256^3$-Hald image} & \multicolumn{3}{c}{Evaluation on natural images}  \\
            objective &  & $\bar{\Delta} E_{M}$$\downarrow$ & $\bar{\Delta} E_{90\%}$$\downarrow$ & PSNR (dB)$\uparrow$  & $\bar{\Delta} E_{M}$$\downarrow$ & $\bar{\Delta} E_{90\%}$$\downarrow$ & PSNR (dB)$\uparrow$ \\
            \hline
            & & & & & & & \\[-7pt]
            \multirow{2}{*}{$L_2$} & Uniform & $2.01{\scriptstyle \pm0.02}$ & $3.93{\scriptstyle \pm0.05}$ &  $39.26{\scriptstyle \pm0.10}$ & $2.44{\scriptstyle \pm0.03}$ & $4.38{\scriptstyle \pm0.06}$ &  $39.68{\scriptstyle \pm0.08}$ \\[1pt] 
             & Natural images & $6.63{\scriptstyle \pm0.12}$ & $16.34{\scriptstyle \pm0.37}$ &  $26.13{\scriptstyle \pm0.18}$ & $1.09{\scriptstyle \pm0.01}$ & $2.19{\scriptstyle \pm0.02}$ & $45.87{\scriptstyle \pm0.21}$ \\[1pt]
            \hline
            & & & & & \\[-7pt]
            \multirow{2}{*}{$\Delta E$} & Uniform & $1.65{\scriptstyle \pm0.02}$ &  $3.13{\scriptstyle \pm0.05}$ &   $36.95{\scriptstyle \pm0.12}$ & $1.92{\scriptstyle \pm0.01}$ &  $3.41{\scriptstyle \pm0.02}$ &  $39.95{\scriptstyle \pm0.15}$ \\[1pt]
            & Natural images & $6.04{\scriptstyle \pm0.21}$ &  $15.07{\scriptstyle \pm0.53}$ &  $25.23{\scriptstyle \pm0.28}$ &  $0.96{\scriptstyle \pm0.02}$ & $1.95{\scriptstyle \pm0.06}$ &  $45.67{\scriptstyle \pm0.19}$  \\[1pt]
        \end{tabular}
    }
    \vspace{0.5cm}
    \caption{This table shows the results of using different color distributions $\mathcal{P}$ and training objectives for training and then evaluating the model (i.e., uniform versus natural images). Results are averaged over five trial runs, each fitting 512 different LUTs using our medium-sized architecture. Training on natural color distribution leads to better performance on natural images and reduced performance when evaluated over the entire color space (i.e., evaluated against Hald images). Using the $\Delta E$ loss function, as expected, reduces color reconstruction errors in terms of $\Delta E$.}
    \label{tab:distribution}
\end{table*}

\subsection{Quantitative Results}
\label{experiment:fitting-hald}

We start with training our models by sampling the LUTs over the entire color space.  We refer to this as uniform sampling as every color in the color space is weighted equally. As previously mentioned, we use a GPU-based interpolation of the LUTs to generate these samples over the LUTs.  

The number of trainable parameters for each model consists of: (1) the core parameters of each $T_i$, and (2) the embedding weights $E_i$s, which grow linearly with the number of LUTs.  With this in mind, we experimented with different numbers of $T_i$s ($D$) of varying depths and widths to pick the right set of hyper-parameters for our network.  We found a hidden structure of $[32, 64, 32]$ neurons for $T_i$s to work best with minimal memory requirements, as the core structure only has $4.3K$ trainable parameters, and embedding each LUT only requires learning an additional 32 parameters. 

We consider four variants of our model (tiny, small, medium, large) with increasing modeling capacity by stacking 1 to 4 $T_i$s together. Figure~\ref{fig:hald-scaling} shows the quality of the color reconstructions for each model variant when trained to embed a varying number LUTs. Smaller models can be used depending on the number of LUTs that need to be embedded. 
Figure~\ref{fig:hald-scaling} also shows that even the smallest model can embed up to 32 LUTs with a low enough $\Delta E$. 

Table~\ref{tab:hald-512} provides a more detailed look at the different models when embedding 512 LUTs. 
Here we report the compression ratios as the ratio of the compressed model checkpoints (including the weights of $T_i$s and $E_i$s) to the file size of the binary LUTs, compressed together in a single archive. See supplemental materials for details.
As expected, the larger models achieve higher fidelity as the number of LUTs increases.
Nonetheless, compared to the average compressed file size of 124.43 MBs for 512 LUTs, even our largest model has a small storage requirement of less than 350 KB.
This represents a $\ge99.7\%$ compression ratio while maintaining a minimal loss of perceivable quality of the reconstructed colors.

\begin{figure}[t]
    \centering
    \includegraphics[width=1\linewidth]{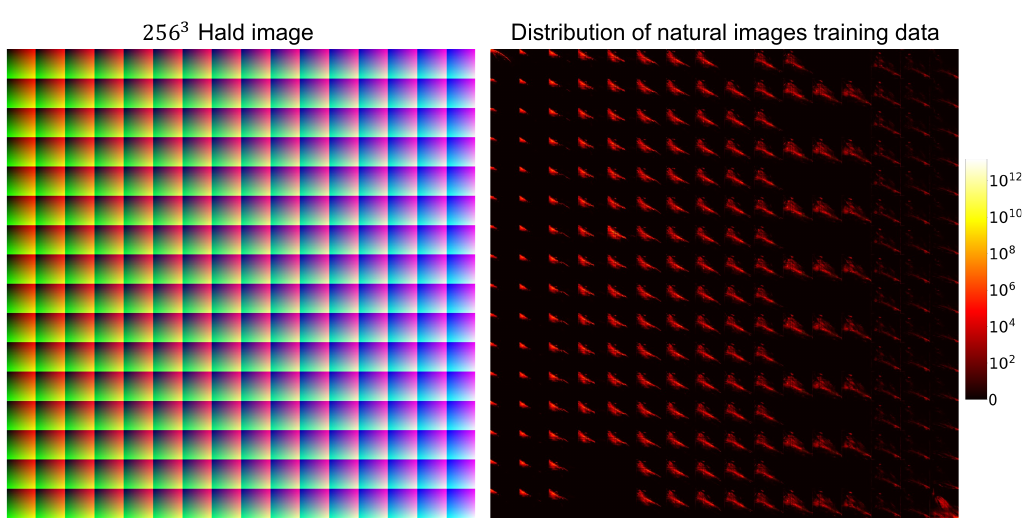}
    \caption{This figure shows a Hald image representing all input colors (left) and
     the distribution of these colors in the 100 Adobe-MIT5K
    images used for training (right).}
    \label{fig:distribution}
\end{figure}

\begin{figure}[t]
    \centering
    \includegraphics[width=0.95\linewidth]{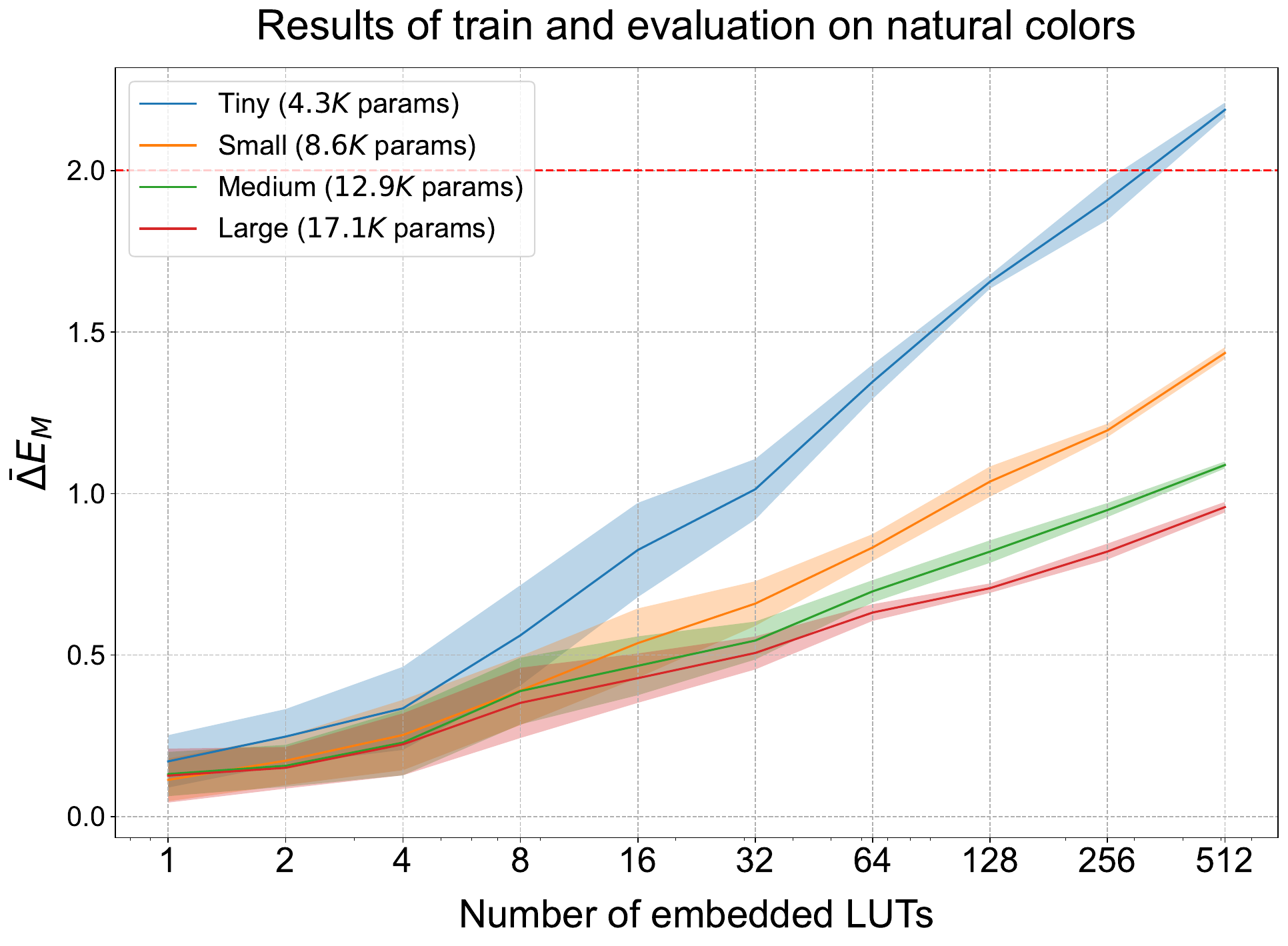}
    \caption{This figure shows the performance of different model sizes when embedding a varying number of LUTs trained and tested on natural images with an $L_2$ training objective. The number of parameters in the model size represents the parameters in the $T_i$ blocks, not counting the $E_i$s.}
    \label{fig:natural_scaling}
\end{figure}

\begin{figure*}[t]
    \centering
    \includegraphics[width=0.85\linewidth]{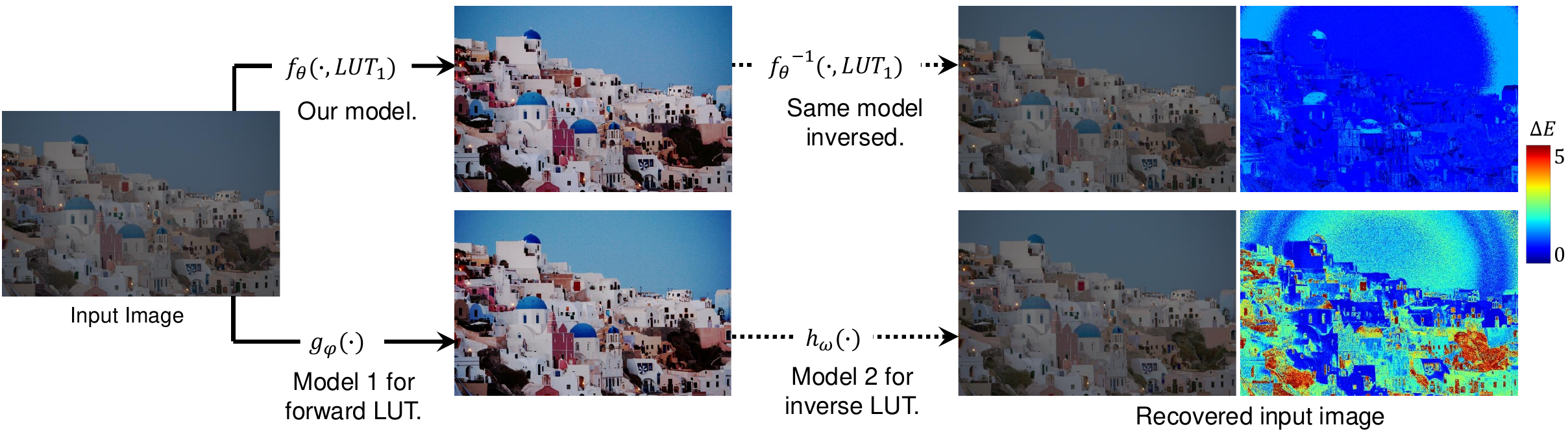}
    \caption{
    This figure compares the inversion accuracy of our modified invertible model versus a non-invertible architecture for estimating the forward and inverse pass through a single LUT. We use a deeper architecture ($D=32$, hidden structure $[16,32,16]$) to embed 32 LUTs, ensuring it remains bijective during training. We train a \textit{small} architecture to estimate the LUT ($g_\psi$) and its inverse ($h_\omega$). Results are shown on an Adobe 5K dataset image processed by reconstructed LUTs, with $\Delta E$ error maps computed per pixel against the ground truth image.
    }
    \label{fig:inversion}
\end{figure*}

\subsection{Alternative Training Approaches}
\label{experiments:distribution}

\subsubsection*{Training using natural images.}~As previously mentioned, the choice of the training color distribution $\mathcal{P}$ plays a role in the overall quality of color reconstructions.
To test this, we randomly select 200 images from the Adobe-MIT5K dataset~\cite{bychkovsky2011learning}, splitting them into a training and testing split of 100 images each.  
Figure~\ref{fig:distribution} shows a visualization of the distribution of the colors in the natural training images as a heat map corresponding to a Hald image.
The distribution heat map reveals that large regions of the color space have a low probability of occurring in the natural images, with approximately $85\%$ of all colors rarely if ever, being observed.

This sparsity suggests that a model trained to focus on those regions of the color space which are more likely to occur in natural images should perform better when evaluated on images.
We repeat our uniform distribution experiments but instead train by using the distribution of colors in the Adobe-MIT5K training split as our $\mathcal{P}$.
We then evaluate the model on the images in the test split. 

Table~\ref{tab:distribution} shows the performance of our medium-sized model trained and tested on different $\mathcal{P}$s to embed 512 LUTs. 
Training uniformly across the entire color space improves generalization across all possible colors. 
However, as seen in Figure~\ref{fig:natural_scaling} and Table~\ref{tab:distribution}, it is evident that knowing the target evaluation color distribution (e.g., when LUTs are consistently applied to natural images) allows for the utilization of specific training distributions $\mathcal{P}$ to enhance the quality of the embedded LUTs.

\subsubsection*{Training with $\Delta E$.}
\label{delta_e_training}

We can also train our network using $\Delta E$ directly as the loss function.  
Table~\ref{tab:distribution} shows that this results in better $\Delta E$ values, but at the cost of lower performance in terms of PSNR.

\subsection{Qualitative Results}
\begin{figure*}[t]
    \begin{center}
    \includegraphics[width=0.95\textwidth]{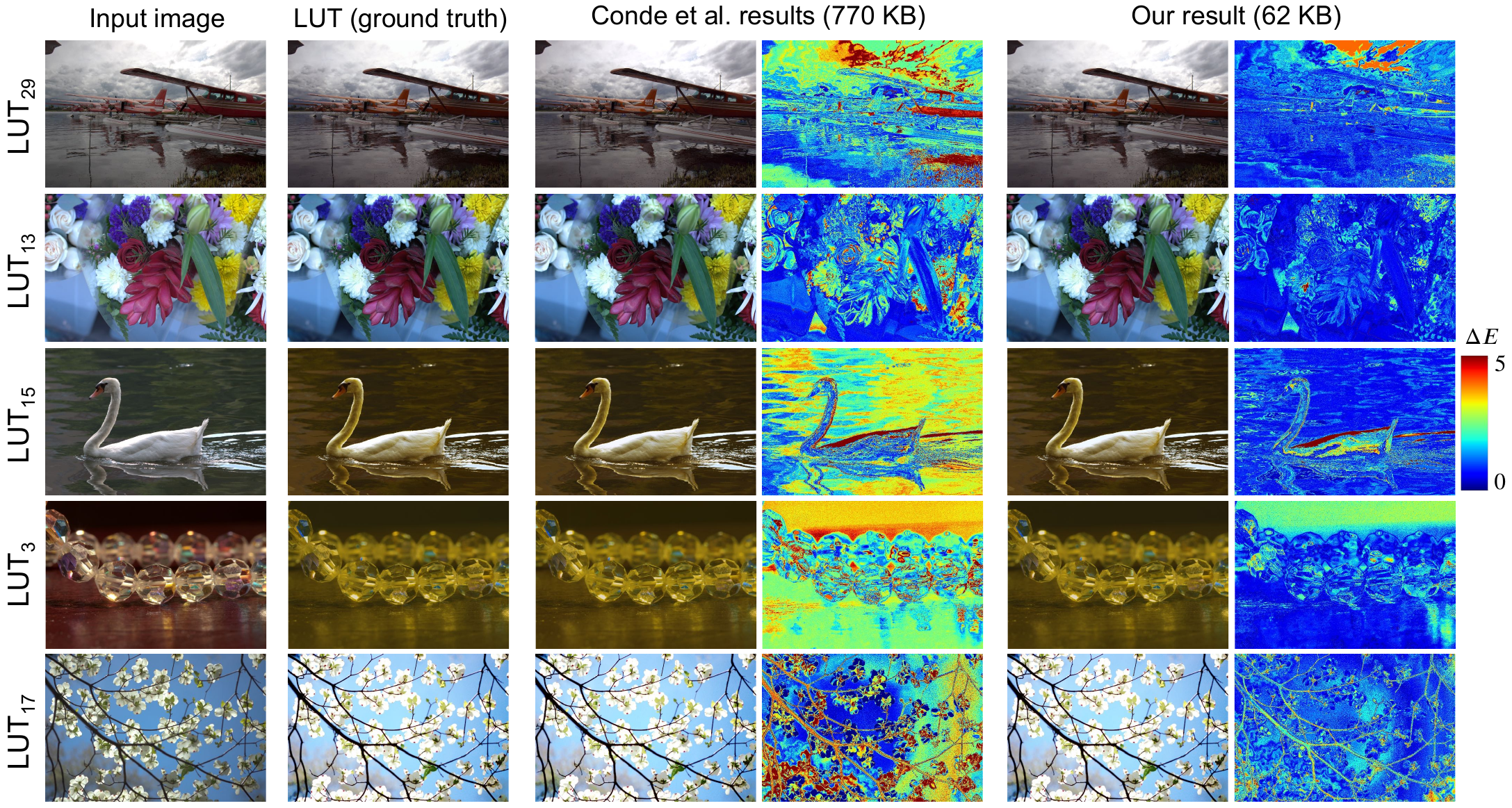}
    \end{center}
    \caption{Comparison of our results with the best-performing model from~\cite{conde2024nilut}. We use our {medium}-sized variant that has embedded 32 LUTs trained on $256^3$ Hald images. Results are shown on images selected from the Adobe 5K dataset~\cite{bychkovsky2011learning} processed by reconstructed LUTs. $\Delta E$ error maps are computed per pixel against the ground truth images that have been processed directly by the corresponding LUT.}
    \label{fig:comparison_marcos}
\end{figure*}

Figure~\ref{fig:comparison_marcos} provides qualitative results on images processed by reconstructed LUTs.
Our results are computed using our medium-sized model with 32 LUT embedding and trained with uniform sampling.
We compared this with the best-performing architecture reported by Conde et al.~\cite{conde2024nilut}. 
Our model can achieve higher fidelity color reconstructions with a significantly smaller model size.  
See supplemental material for further details on these experiments and architectural ablation studies.

\begin{figure*}[ht]
    \begin{center}
    \includegraphics[width=0.95\textwidth]{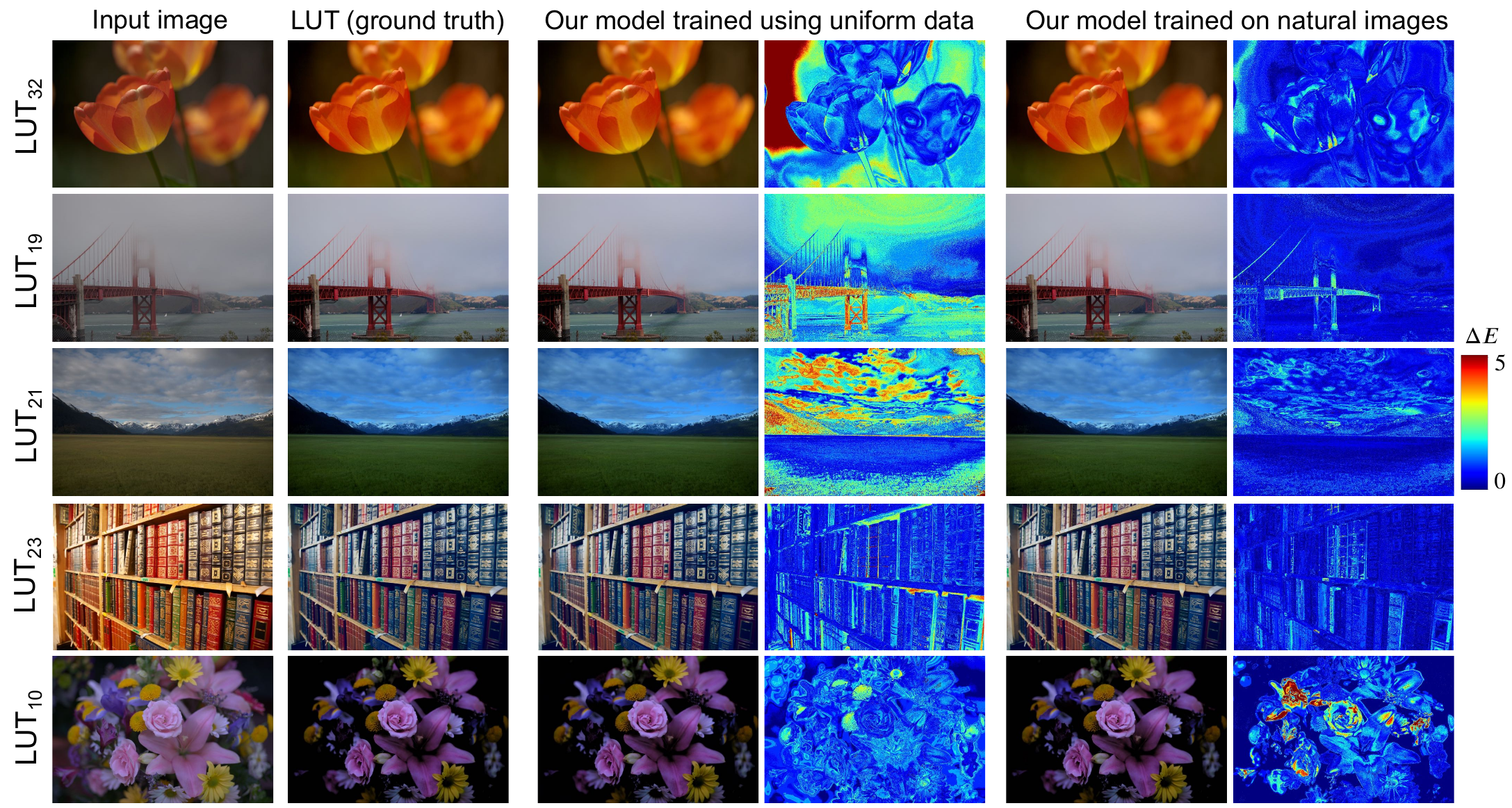}
    \end{center}
    \caption{Comparison of the result of training directly on 100 natural color images versus uniform training. The medium-sized variant of our network is trained to embed 32 LUTs. The images shown are selected from the testing image split. Training on natural images provides a notable improvement when applied to natural images.
    }
    \label{fig:qualitative_natural}
\end{figure*}

Figure~\ref{fig:qualitative_natural} shows qualitative results on our model trained using uniform sampling and natural image sampling.  
As the quantitative experiments also indicate, training our models on natural images improves performance when the reconstructed LUTs are applied to natural images.  
Nevertheless, uniform training might be preferred when consistency of predictions is required, as the model trained on a sparse distribution might fail to faithfully reconstruct colors that appear infrequently in the training distribution. 
For example, in Figure~\ref{fig:qualitative_natural} (last row), the purple-colored flowers have higher $\Delta E$ than those processed by the uniformly trained model.

\subsection{LUT Inversion}
\label{experiment:inverse}
We designed our model to be initialized near the space of bijective transformations.
Here we examine the effect of keeping the network strictly bijective.
We use spectral normalization~\cite{miyato2018spectral, gouk2021regularisation} to normalize the weights of each residual transformation $T_i$ by its largest singular value, enforcing a Lipschitz constant of $0.97$.
This restriction, along with our choice of activation function, forces the model to remain bijective during training, which allows the network to be inverted \cite{behrmann2019invertible} with a fixed-point iteration algorithm. As a result, we can compute an inverse color LUT by reversing the order of computations in the network and inverting each transformation. See supplemental materials for details.

Restricting the architecture to be bijective may slightly decrease its modeling capacity, and we find more depth is needed to approximate the LUTs comparably. For more details, see the supplemental material.

We visualize the results of our LUT inversion mechanism on a single image and LUT in Figure \ref{fig:inversion}.
For a point of comparison, we fit our {small}-sized architecture on the LUT applied (referred to as LUT$_1$) as follows.
First, we fit the LUT as normal ($g_\psi$).
Next, we also fit the LUT but swap inputs and outputs to estimate its inverse ($h_\omega$).
As seen in Figure \ref{fig:inversion}, the invertible architecture works well at inverting the LUT, effectively up to the numerical precision of the fixed-point iteration.
In contrast, attempting to directly estimate the LUT and its inverse separately results in significant errors.

\section{Concluding Remarks}
\label{sec:discussion}
This paper has introduced a network architecture designed to efficiently encode 3D color lookup tables (LUTs) into a single compact representation. 
Our proposed model achieves this with a minimal storage footprint, consuming less than 0.25 MB.
The reconstruction capability of the model extends to 512 LUTs, introducing only minor color distortion ($\bar{\Delta}E_M$ $\leq$ 2.0) across the entire color space.
Furthermore, we demonstrate that the network has the ability to weight LUT colors, yielding additional quality improvements, particularly evident in natural image colors with $\bar{\Delta}E_M$ $\leq1.5$.
Our network architecture is also able to accommodate bijective encoding, enabling the production of invertible LUTs and facilitating reverse color processing.

\appendix

\makesupplementtitle{Supplementary Material}

\section{Data}
\label{sec:supp_data}
\begin{figure*}[h]
    \begin{center}
        \includegraphics[width=0.95\linewidth]{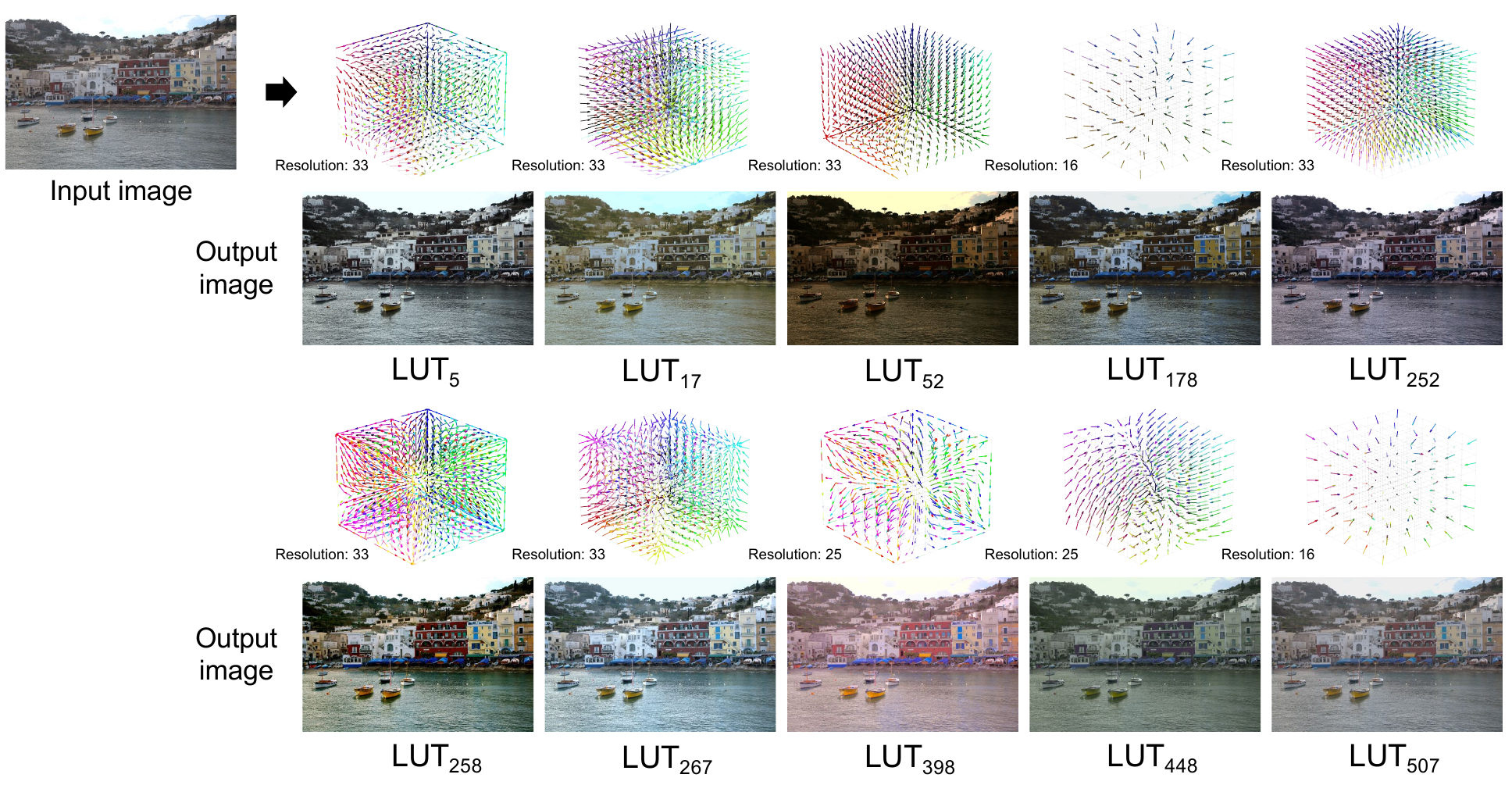}
    \end{center}
    \caption{This figure shows a random selection of the LUTs in our collection. Our LUTs exhibit different color manipulations and are at different lattice resolutions.  Needle maps are sub-sampled by a factor of three for better visualization.}
    \label{fig:supp-luts}
\end{figure*}

As discussed in the main text,
we use a collection of publicly available LUTs under a Creative Common License. Figure \ref{fig:supp-luts} shows example LUTs from our collection, which covers a diverse set of color transformations used in practice.

\section{Experimental Setup \& Details}
\label{sec:supp_training}
\subsection{Training details}
The main architectural building block ($T_i$) in our network has a hidden structure of $[32, 64, 32]$ neurons. At every optimization step, we use a batch of 2048 colors. We target all the embedded LUTs (at most 512 LUTs) for a total of 30760 iterations. We use Adam \cite{kingma2014adam} with a learning rate of 0.04. We decrease the learning rate by half at every 2560 iterations.

We assume all colors are encoded with 8-bits per RGB channel.
Starting with color $c$, we normalize each {of its channel} values by:
\begin{align}
    \bar{c} =& 2a(\frac{c}{255} - \frac{1}{2}),
\end{align}
such that $\bar{c} \in [-a,a]$ which bounds the input space for training.
Since the model's output is passed through $\tanh$, the gradient of every parameter before the $\tanh$ will be multiplied by $\tanh'$ in the backward pass.
By setting  $a=0.83$ in all our experiments we ensure that the values passed through $\tanh$ will fall into a region with large enough gradients ($\ge\frac{1}{2}$), mitigating any possible issues with vanishing gradients.
This normalization is inverted for evaluating and visualizing results, and the values are clipped to be in $[0,1]$ and scaled and quantized back to 8-bits.

As mentioned in our description of our proposed neural architecture,
we want our models to be initialized close to an identity and a bijective function. To do this, we initialize our network weights and biases to have small values by taking PyTorch's \cite{paszke2019pytorch} stock initialization of the weights and biases and dividing them by $100$. We observed that when we initialize our biases near-zero values, they change little during training, and the network incurs negligible loss in the quality of color reconstructions if they are dropped entirely. As a result, in all our experiments, we turn off the biases for all the linear layers to further save on memory and compute resources.

\subsection{Architectural Ablation}
We perform ablations to test the three major contributing factors to our model's ability to embed LUTs efficiently.  These factors are as follows:
 \begin{enumerate}
     \item \textbf{Sampling}~Training with a sampling-based approach instead of using actual Hald-images to circumvent  computational costs when embedding a large number of LUTs,
     \item \textbf{Inductive-biases} Careful initialization of the model near a bijective and an identity function, in addition to the use of specifically curated normalization steps,
     \item \textbf{Model architecture} Our model was designed with multiple residual connections, in contrast to Conde et al.  \cite{conde2024nilut}, which used a wider network with a single residual connection in its predictive path.
 \end{enumerate}

First, we perform an ablation study on how a different architecture from \cite{conde2024nilut} that has a similar memory footprint as our medium-sized model performs if it was trained using our training framework. We pick the MLP-Res with $N=3$ and $L=64$ from \cite{conde2024nilut} as our architectural baseline, and we compare its performance against our medium-sized model. 

We train both models to embed 32 LUTs simultaneously using our sample-based algorithm. However, to study the effects of using alternative architectures, we train the MLP-Res model in two ways. First, we use the sampling-based training method with an adjusted learning rate of 0.002 and adhere to all other training procedures described in \cite{conde2024nilut}. Second, we modify the model with our recommended initialization and normalization and train it using our sample-based approach. Table \ref{tab:ablation} summarizes the results for each training method in comparison to the performance of our medium-sized model. It is clear that both the architecture and the inductive biases significantly affect the quality of color reconstructions.

\begin{table*}[t]
    \centering
    \resizebox{0.95\textwidth}{!}{%
        \begin{tabular}{c:c|ccc|ccc}
            \multirow{2}{*}{Model Size} & \multirow{2}{*}{\#Parameters $\downarrow$} & \multicolumn{3}{c}{Evaluation on a $256^3$-Hald image} & \multicolumn{3}{c}{Evaluation on natural images}  \\ & & $\bar{\Delta} E_{M}$$\downarrow$ & $\bar{\Delta} E_{90\%}$$\downarrow$ & PSNR (dB)$\uparrow$ & $\bar{\Delta} E_{M}$$\downarrow$ & $\bar{\Delta} E_{90\%}$$\downarrow$ & PSNR (dB)$\uparrow$ \\
            \hline
            & & & & & & \\[-4pt]
            MLP-Res \cite{conde2024nilut} & 12.9K & $1.54{\scriptstyle \pm0.19}$ & $3.17{\scriptstyle \pm0.69}$ & $41.79{\scriptstyle \pm1.01}$ & $1.83{\scriptstyle \pm0.24}$ & $3.42{\scriptstyle \pm0.58}$ & $42.23{\scriptstyle \pm0.90}$\\[3pt]
            W/ inductive-biases & 12.9K & $1.16{\scriptstyle \pm0.32}$ & $2.26{\scriptstyle \pm0.64}$ & $44.01{\scriptstyle \pm2.16}$ & $1.41{\scriptstyle \pm0.34}$ & $2.59{\scriptstyle \pm0.61}$ & $44.09{\scriptstyle \pm1.96}$\\[3pt]
            Ours & 12.9K & $0.83{\scriptstyle \pm0.09}$ & $1.63{\scriptstyle \pm0.19}$ & $46.60{\scriptstyle \pm1.05}$ & $1.07{\scriptstyle \pm0.12}$ & $2.01{\scriptstyle \pm0.22}$ & $46.31{\scriptstyle \pm1.02}$\\
        \end{tabular}
    }
    \vspace{0.1cm}
    \caption{This table shows an ablation performed to verify the effectiveness of our network architecture and training framework. The first row is an MLP-Res from Conde et al.~\cite{conde2024nilut} with 12.9K parameters, the second row is the same model but initialized using our proposed method and trained using our normalization, and the third row is our proposed method with our {medium}-sized architecture. Results are averaged over 5 trial runs trained uniformly to embed 32 LUTs, with each trial using a different set of LUTs.}
    \label{tab:ablation}
\end{table*}

\begin{figure}[t]
    \centering
    \includegraphics[width=\columnwidth]{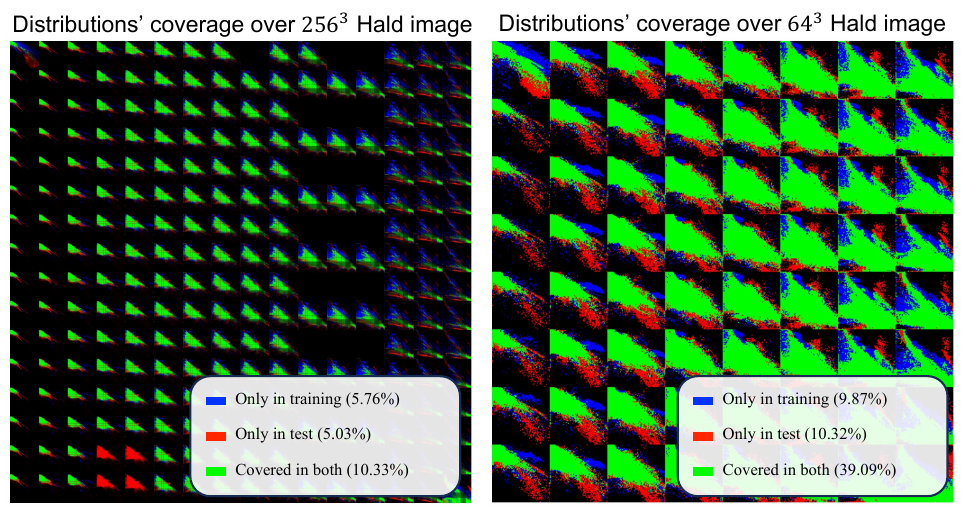}
    \caption{Comparing the color distribution of our training and test natural images splits, we see more overlap as we increase the binning size and approach the voxel dimensions of typical LUTs in our collection.}
    \label{fig:distribution}
\end{figure}

\begin{table}[t]
    \centering
    \resizebox{1\linewidth}{!}{%
        \begin{tabular}{c|ccc}
        Evaluation method & $\bar{\Delta} E_{M}$$\downarrow$ & SSIM $\uparrow$ & PSNR (dB)$\uparrow$ \\[3pt]
        \hline
         & & & \\[-4pt]
        Evaluation on 100 random images  & $1.3809$ & $0.9790$ &  $44.2853$\\[4pt]
        Evaluation on all the MIT5K dataset  & $1.4025$ & $0.9783$ &  $44.1310$ \\[3pt]
        \end{tabular}
    }
    \vspace{0.1cm}
    \caption{The effect of increasing the size of the evaluation dataset. Results are based on our {medium}-sized model trained uniformly to fit 32 LUTs. As seen, there is minimal difference in terms of the measured numbers, therefore we opt for using 100 random images in our experiments for faster evaluation times.}
    \label{tab:mit5k}
\end{table}

\begin{table*}[t]
    \centering
    \resizebox{0.95\linewidth}{!}{%
        {\large
        \begin{tabular}{l:c|c:c:c:c}
        \multicolumn{1}{c}{\multirow{2}{*}{Starting LUTs to compress}} & Average file size & \multicolumn{4}{c}{Average compression ratio (\%) $\uparrow$}\\
        & of LUTs (MB) & \textit{tiny}-model & \textit{small}-model & \textit{medium}-model & \textit{large}-model \\
        \hline
        & & & & & \\[-4pt]
        512 LUTs in \texttt{.cube} format & 381.774 & $99.979$ & $99.959$ & $99.938$ & $99.918$ \\[8pt]
        \hdashline
        & & & & & \\[-4pt]
        512 LUTs in \texttt{.cube} compressed in a single \texttt{.zip} archive  & 131.335 & $99.940$ & $99.880$ & $99.821$ & $99.761$  \\[8pt]
        \hdashline
        & & & & & \\[-4pt]
        512 LUTs in \texttt{.npy} 32-bit binary format  & 184.844 &  $99.957$ & $99.915$ & $99.873$ & $99.830$ \\[8pt]
        \hdashline
        & & & & & \\[-4pt]
        512 LUTs in \texttt{.npy} 32-bit binary format which is then  & \multirow{2}{*}{124.787} &\multirow{2}{*}{$99.957$} & \multirow{2}{*}{$99.915$} & \multirow{2}{*}{$99.873$} & \multirow{2}{*}{$99.830$} \\
        compressed in a single \texttt{.zip} archive & & & & & \\[8pt]
        \hdashline
        & & & & & \\[-4pt]
        512 LUTs in \texttt{.npz} 32-bit compressed binary format  & 124.803 &  $99.937$ & $99.874$ & $99.812$ & $99.749$  \\[8pt]
        \hdashline
        & & & & & \\[-4pt]
        512 LUTs in \texttt{.npz} 32-bit compressed binary format & \multirow{2}{*}{124.246} & \multirow{2}{*}{$99.936$} & \multirow{2}{*}{$99.874$} & \multirow{2}{*}{$99.811$} & \multirow{2}{*}{$99.748$} \\
        which is then compressed in a single \texttt{.zip} archive & & & &   \\
        \end{tabular}    
    }
    }
    \vspace{0.1cm}
    \caption{This table provides alternative compression ratio computations based on other file formats.}
    \label{tab:compression}
\end{table*}

\begin{table*}[h]
    \centering
    \resizebox{0.8\linewidth}{!}{%
        {\large
        \begin{tabular}{l|cc|c}
        \multicolumn{1}{c|}{Task} & CPU$_1$ {\footnotesize (ms)} & GPU$_1$ {\footnotesize (ms)} & GPU$_2$ {\footnotesize (ms)}\\[5pt]
        \hline
        & & \\[-5pt]
        Reconstructing a single LUT at 7$\times$7$\times$7 resolution &  $0.97{\scriptstyle \pm0.07}$ &  $6.83{\scriptstyle \pm0.39}$ &  $1.21{\scriptstyle \pm0.04}$ \\[4pt]
        Reconstructing a single LUT at 11$\times$11$\times$11 resolution &  $2.48{\scriptstyle \pm0.23}$ &  $7.11{\scriptstyle \pm0.40}$ &  $1.27{\scriptstyle \pm0.02}$ \\[4pt]
        Reconstructing a single LUT at 33$\times$33$\times$33 resolution &  $28.50{\scriptstyle \pm0.18}$ &  $16.20{\scriptstyle \pm0.88}$ &  $1.64{\scriptstyle \pm0.05}$ \\[4pt]
        Reconstructing a single LUT at 65$\times$65$\times$65 resolution &   $159{\scriptstyle \pm9.64}$ &  $54{\scriptstyle \pm8.27}$&  $4.11{\scriptstyle \pm0.03}$ \\[3pt]
        \hline
        & & \\[-4pt]
        Transforming a 12MP
        image pixel by pixel through our model  & $7460{\scriptstyle \pm31.2}$ & $2100{\scriptstyle \pm193}$ & $239{\scriptstyle \pm0.06}$ \\
        \end{tabular}    
        }
    }
    \vspace{0.1cm}
    \caption{Run-times for our medium-sized model, using (1) Apple M1 Pro (CPU$_1$, GPU$_1$), and (2) Nvidia RTX4090 (GPU$_2$). The mean (standard deviation) is computed over a hundred trials. It is important to note that no runtime optimization is applied to the implementation of our method in this experiment, and further enhancement of these results could be expected with hardware specific, or better optimized implementations.}
    \label{tab:runtime}
\end{table*}

\subsection{Evaluations}
To estimate the reconstruction quality of colors in natural images, we evaluate our models against the colors found in images from the Adobe-MIT 5K dataset~\cite{bychkovsky2011learning}. Instead of evaluating our results on all of the images in this dataset, we use 100 randomly selected images for faster evaluations. This choice was validated by comparing against evaluation with a larger number of images, which did not significantly alter our results. Table~\ref{tab:mit5k} demonstrates the minimal difference between evaluating using 100 images versus the entire MIT5k dataset~\cite{bychkovsky2011learning}. Additionally, regarding other metrics that consider spatial correlations and structures in images, namely SSIM~\cite{wang2004image}, we did not include these metrics in our main experiments because LUTs perform per-pixel transformations which tend to minimally affect the spatial structures of images rendering such metrics redundant.

Using a uniform training distribution puts equal importance on all colors, which is ideal for applications that either require high reconstruction accuracy over the whole color space or have an unknown or unusual color distribution.
However, training on a narrower color distribution (e.g., one based on natural images) allows the model to sacrifice the accuracy for certain colors that are more relevant to the downstream task over other colors in the input space.
Thus, when such a model is evaluated on the full color gamut (e.g., a Hald image) we expect to see degraded worst case performance for the benefit of improved reconstruction qualities on in-distribution colors.

To further justify the applicability of our sample-based training algorithm and evaluation procedure, it is important to note that LUTs are generally smooth. Therefore, in order to embed a typical LUT we do not need to visit all the possible input colors specifically during training; instead, as long as our samples cover all the input-output voxels of the LUT, we expect to capture its behaviour. In our finite-sample training procedure we expect to visit each voxel for a high resolution 64$^3$ LUT over 200 times ($\frac{2048\times30760}{64^3}$) on average which as seen in our experiments, is practically sufficient to capture the behaviour of the embedded LUTs. Similarly, when thinking about evaluating the performance of the model on custom distributions, as long as the training and test color distributions cover the same proximity as the input-output voxels, we expect our model to generalize. We validate that this is the case for our experiments that target colors in natural images by comparing the coverage of our training and test color distributions as seen in Figure~\ref{fig:distribution}.

\subsection{Compression Ratios}
Several standardized formats are used to store and distribute LUTs. Here, we consider alternative file formats and provide additional compression ratios in addition to those provided in the main text for completeness.

A standard \texttt{.cube} text format is commonly used for distributing LUTs online. Alternatively, you may also find transformed Hald images in a \texttt{.png} format as a representation for a LUT, which at a 256$\times$256$\times$256 resolution can reach a file size of well above 4 MB for each LUT. However, it is unlikely for such a format to be used in resource-constrained applications. Instead, in such restrictive conditions, using a simple binarized representation of the LUT lattice is a common practice to store LUTs more efficiently.   

At the same time, there are multiple ways that a neural network could be stored and deployed. In this work, we consider a simple compressed binary version of our model weights (including $T_i$s and $E_i$s) as 32-bit floating-point numbers represented in the \texttt{.npz} file format to compute our model storage requirements and later calculate our compression ratios. \texttt{.npz} is a commonly used off-the-shelf compressed file format used for the storage of numerical tensors provided in the NumPy package~\cite{harris2020array}. In our main text, we compare the file size of our model checkpoints in \texttt{.npz} with the average file size of a single zipped archive of all \texttt{.npz} representations of the embedded LUTs in the model to get our compression ratios.  Table~\ref{tab:compression} provides additional compression ratios based on other storage formats.

\subsection{Computational Cost}
We envision our work will be most relevant to applications running on resource constrained devices, to construct individual 3D LUTs (e.g., 33$\times$33$\times$33) on demand, to be temporarily kept in memory for as long as necessary. Our model represents hundreds of LUTs, eliminating the requirement to manage or swap between many different LUTs in a single application. Alternatively, our method can also process images directly by applying the model to each pixel. Nonetheless, we emphasize that extracting and using the LUT is our intended use case for our method as employing specialized LUT code or hardware is significantly more efficient when it comes to processing images. Table~\ref{tab:runtime} shows that the computational cost of extracting LUTs is very reasonable.  Applying our method to a full-sized image would not be encouraged (and is not the intended use of our method).

\section{LUT Bijectivity (Inversion)}
\label{sec:supp_bijective}

As discussed in the main text, if every residual component of a ResNet is restricted to be contractive, the whole network is bijective (invertible) by the Banach fixed-point theorem \cite{behrmann2019invertible}. Therefore, to limit our model to stay in the bijective function-space, we guarantee the contractiveness of each $T_i$ in our model.

$T_i$s consist of a stack of linear layers with $\lipswish$ non-linearities, and the overall lipschitz constant of $T_i$ depends on the product of the derivatives of the non-linearities and the largest singular values of the weights for its linear layers. $\lipswish$ has bounded derivatives $|\frac{\mathrm{d}}{\mathrm{d}x}\mathrm{\lipswish}|\le1$, hence the contractiveness of $T_i$ will solely depend on the contractiveness of its linear layers.

We limit the contractiveness of the linear weights ($W_{i,j}$) using Spectral Normalization \cite{kobyzev2020normalizing,gouk2021regularisation}:
\begin{equation}
    \Tilde{W}_{i,j} = c_{\text{coefficient}} \cdot \frac{W_{i,j}}{\Tilde{\sigma}_{\max}(W_{i,j})},
\end{equation}
where $\Tilde{\sigma}_{\max}(\cdot)$ indicates the magnitude of the largest singular values of a given matrix, and  $ \Tilde{W}_{i,j}$ denotes the resulting normalized weights used for the $j$-th hidden layer in $T_i$. $c_{\text{coefficient}}$ is an arbitrary number used for controlling the contractiveness of each $W_{i,j}$. We use $\Tilde{T}_i$ to differentiate restricted $T_i$s from regular $T_i$s.

Our model computes the corresponding outputs for given inputs $x$ as:
\begin{equation}
    f_\theta(x, \idx) = \tanh(T_D\circ \cdots \circ T_1(\tanh^{-1}(x), \idx)),
\end{equation} 
therefore the inverse path could be written as:
\begin{equation}
    f^{-1}_\theta(x, \idx) = \tanh(\Tilde{T}^{-1}_1\circ \cdots \circ \Tilde{T}^{-1}_D(\tanh^{-1}(x), \idx)),
\end{equation} 
where $\Tilde{T}^{-1}_i$ is approximated with fixed-point iteration \cite{behrmann2019invertible} as described in Algorithm \ref{algo:inverse}.

\begin{algorithm}
    \begin{algorithmic}[1]
    \State \textbf{Input:} output from the residual component $y$, the residual component $\Tilde{T}_i$, number of fixed-point iterations $n$
    \State \textbf{Init:} $x_0 := y$
    \For{$j = 0, \ldots, n$}
       \State $x_{j+1} := y - \Tilde{T}_i(x_j)$
    \EndFor
    \end{algorithmic}
    \caption{Approximation of $\Tilde{T}^{-1}_i$s via fixed-point iteration.}
    \label{algo:inverse}
\end{algorithm}

\subsection{Inversion Experiment Details}
As mentioned in the main text, the restricted $\Tilde{T}_i$s naturally have less modelling capacity to predict larger residual values, therefore, a restricted model would need a larger number of $\Tilde{T}_i$s and learnable parameters to achieve comparable performance to a non-restricted network. To achieve similar results as our large-sized model, we chose an architecture of 32 $\Tilde{T}_i$s with $[16, 32, 16]$ hidden neurons (with almost twice the memory footprint). We use $c_{\text{coefficient}}=0.97$ and  $n=5000$ fixed-iteration steps. Every other detail for the training is exactly as it was for unrestricted models.

\section{Applications and Future Work}
\begin{figure}[h]

    \begin{center}
        \includegraphics[width=\linewidth]{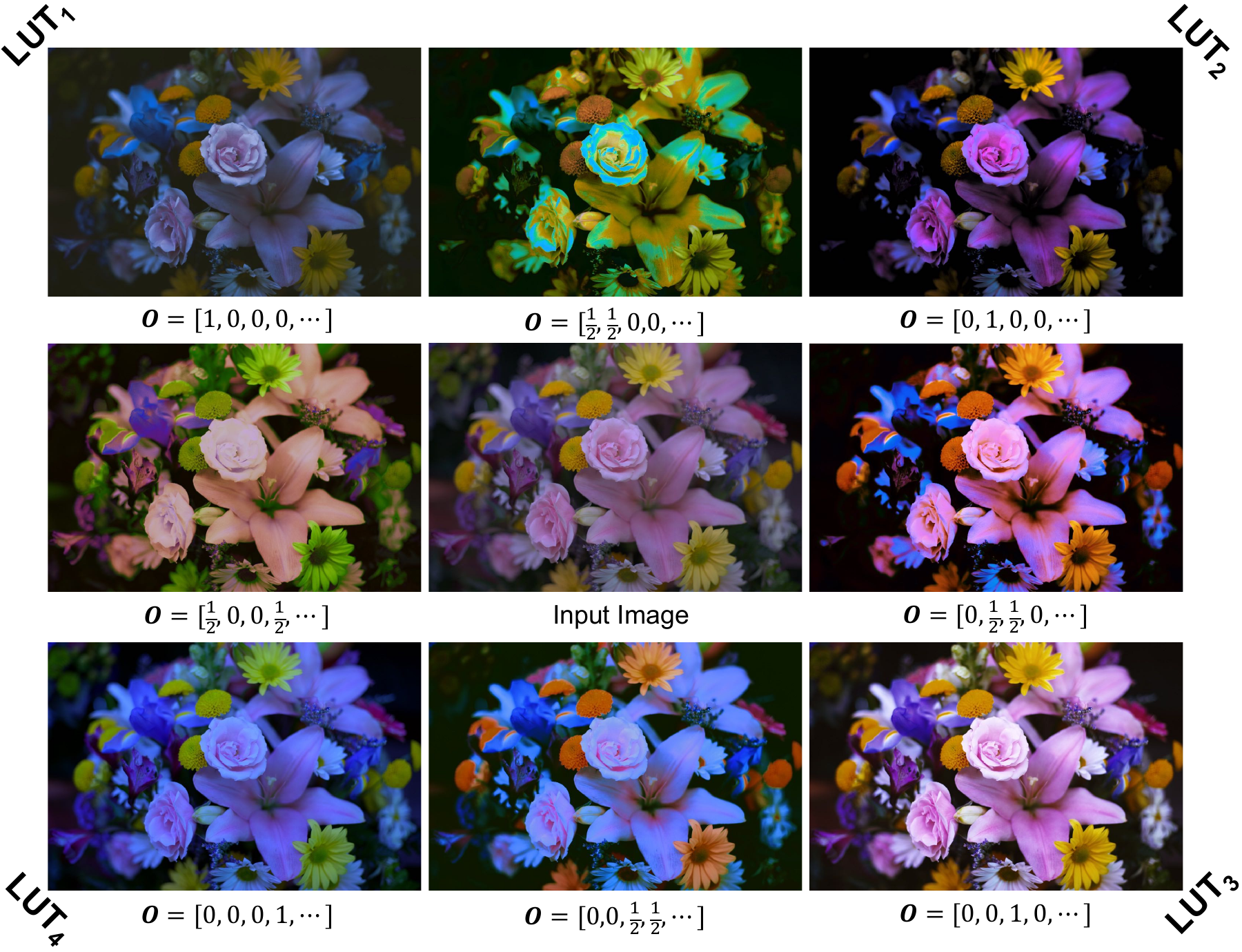}
    \end{center}
    
    \caption{This figure visualizes the result of transforming a sample image (center) drawn from the MIT5K dataset \cite{bychkovsky2011learning} with arbitrary embedded LUTs extracted from our medium-sized model trained uniformly with an MSE objective to embed 512 LUTs. The four corner images show the original image transformed with the reconstructed embedded LUTs (extracted with one-hot encoded indices $\idx$), and the other four images in between the corners show the same image transformed with a blended novel LUT that sits between the two LUTs represented in each corner (extracted with fractional indices $\idx$).}
    \label{fig:blending}
\end{figure}

Implicit LUT representations open up possibilities for lots of new applications well beyond the use cases of 3D color LUTs. Our manuscript focuses on embedding LUTs efficiently, but also demonstrates invertible color transformations as an example for a direct use case of our approach. Similar to \cite{conde2024nilut}, our method supports smooth blending between LUTs by using fractional input indices instead of one-hot encoded indices $\idx$s, specifying the desired weights for embedded LUTs to mix in during color reconstruction as seen in Figure~\ref{fig:blending}. 
We hope this work inspires further research in the applications of neural embedding of LUTs.

\begin{figure*}[t]
    \begin{center}
    \includegraphics[width=0.85\linewidth]{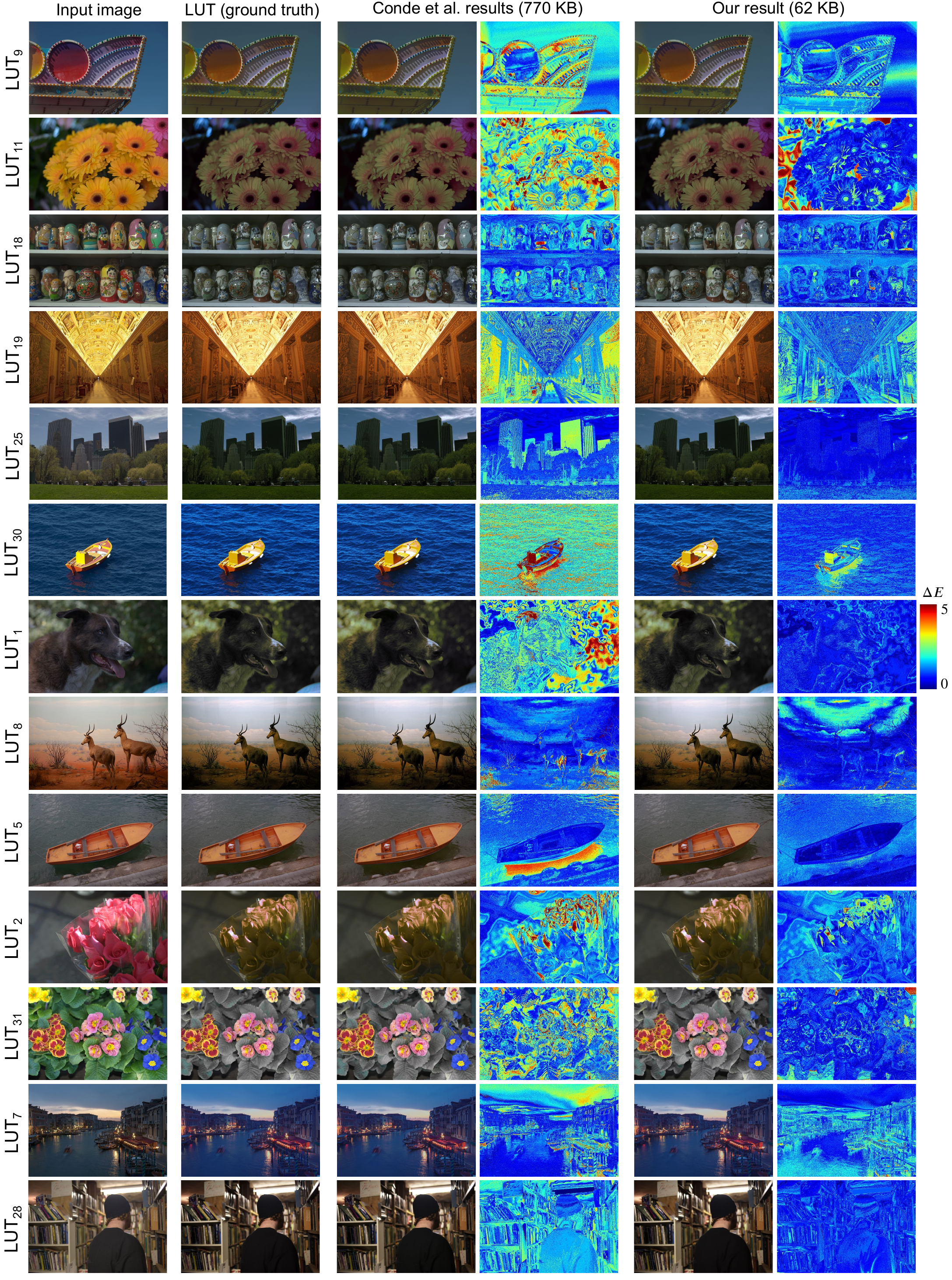}
    \end{center}
    \caption{This figure provides additional examples for Figure~\ref{fig:comparison_marcos} in the main text,
    with different LUTs, comparing the result of our {medium}-sized model with the best-performing model from~\cite{conde2024nilut}. Both models are trained uniformly to embed 32 LUTs. Images are selected from the Adobe 5K dataset~\cite{bychkovsky2011learning} processed by reconstructed LUTs.  $\Delta E$ error maps are computed per pixel against the ground truth images processed directly by the corresponding LUT.}
    \label{fig:supp-comparison_marcos}
\end{figure*}

\begin{figure*}[t]
    \begin{center}
        \includegraphics[width=0.85\linewidth]{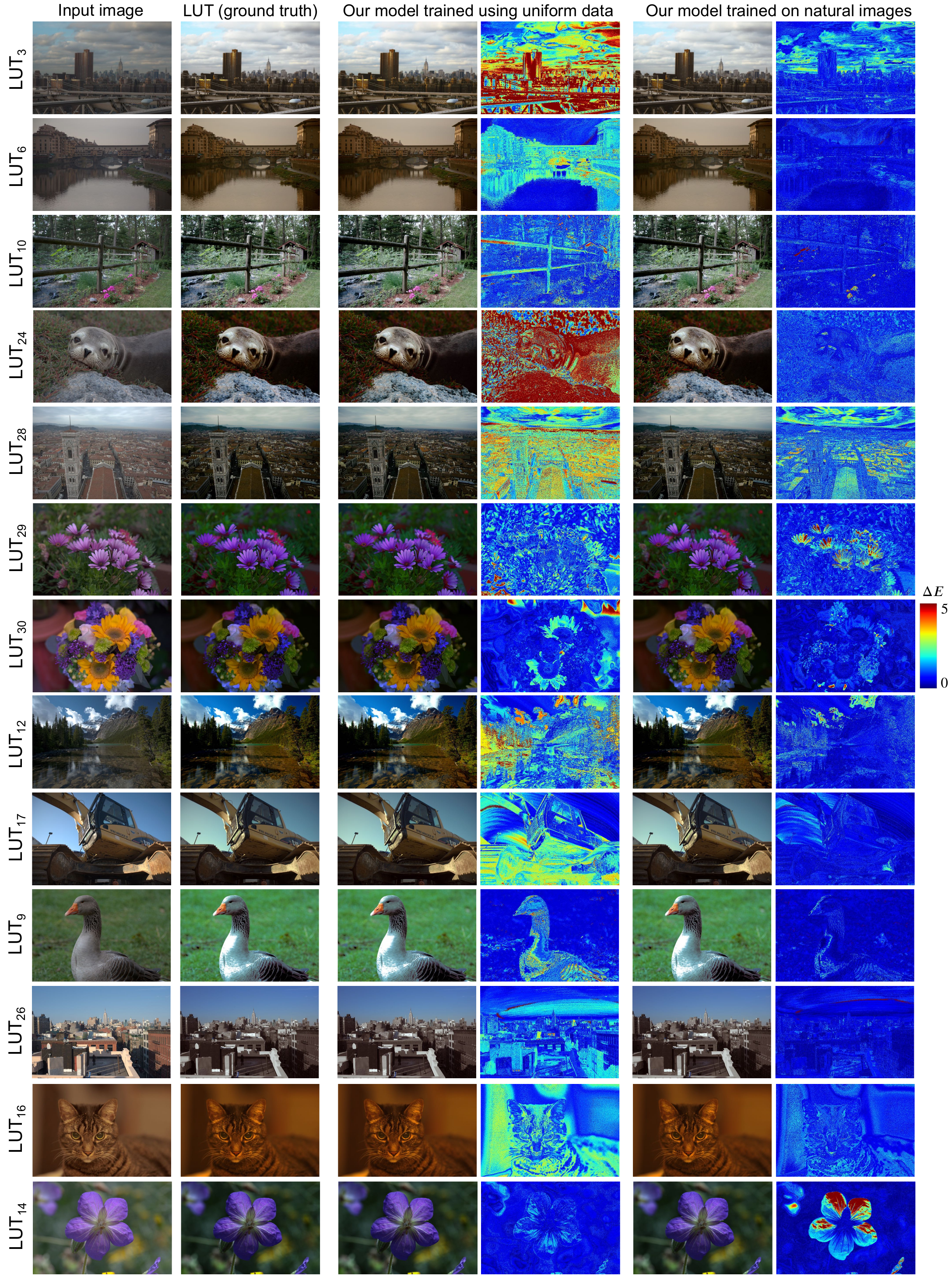}
    \end{center}
    \caption{This figure provides additional examples for Figure~\ref{fig:qualitative_natural} in the main text,
    with different LUTs. Here the result of training directly on 100 natural images versus uniform training is compared. For both distributions, our {medium}-sized model is trained to embed 32 LUTs.  The images shown are selected from the Adobe 5K dataset~\cite{bychkovsky2011learning} (outside of the training image split).  Training on natural images provides a notable improvement when applied to natural images, except for rare colors such as the exhotic purples in the flowers.
    }
    \label{fig:supp-qualitative_natural}
\end{figure*}

\FloatBarrier
{

}

\end{document}